# The Big Data Bootstrap

**Ariel Kleiner**                                                 AKLEINER@CS.BERKELEY.EDU
**Ameet Talwalkar**                                                  AMEET@CS.BERKELEY.EDU
**Purnamrita Sarkar**                                             PSARKAR@CS.BERKELEY.EDU
Computer Science Division, University of California, Berkeley, CA 94720, USA

**Michael I. Jordan**                                             JORDAN@CS.BERKELEY.EDU
Computer Science Division and Department of Statistics, University of California, Berkeley, CA 94720, USA

## Abstract

The bootstrap provides a simple and powerful means of assessing the quality of estimators. However, in settings involving large datasets, the computation of bootstrap-based quantities can be prohibitively demanding. As an alternative, we present the Bag of Little Bootstraps (BLB), a new procedure which incorporates features of both the bootstrap and subsampling to obtain a robust, computationally efficient means of assessing estimator quality. BLB is well suited to modern parallel and distributed computing architectures and retains the generic applicability, statistical efficiency, and favorable theoretical properties of the bootstrap. We provide the results of an extensive empirical and theoretical investigation of BLB's behavior, including a study of its statistical correctness, its large-scale implementation and performance, selection of hyperparameters, and performance on real data.

## 1. Introduction

Assessing the quality of estimates based upon finite data is a task of fundamental importance in data analysis. For example, when estimating a vector of model parameters given a training dataset, it is useful to be able to quantify the uncertainty in that estimate (e.g., via a confidence region), its bias, and its risk. Such quality assessments provide far more information than a simple point estimate itself and can be used to improve human interpretation of inferential outputs, per-form hypothesis testing, do bias correction, make more efficient use of available resources (e.g., by ceasing to process data when the confidence region is sufficiently small), perform active learning, and do feature selection, among many more potential uses.

Accurate assessment of estimate quality has been a longstanding concern in statistics. A great deal of classical work in this vein has proceeded via asymptotic analysis, which relies on deep study of particular classes of estimators in particular settings (Politis et al., 1999). While this approach ensures asymptotic correctness and allows analytic computation, its applicability is limited to cases in which the relevant asymptotic analysis is tractable and has actually been performed. In contrast, recent decades have seen greater focus on more automatic methods, which generally require significantly less analysis, at the expense of doing more computation. The bootstrap (Efron, 1979; Efron & Tibshirani, 1993) is perhaps the best known and most widely used among these methods, due to its simplicity, generic applicability, and automatic nature. Efforts to ameliorate statistical shortcomings of the bootstrap in turn led to the development of related methods such as the $m$ out of $n$ bootstrap and subsampling (Bickel et al., 1997; Politis et al., 1999).

Despite the computational demands of this more automatic methodology, advancements have been driven primarily by a desire to remedy shortcomings in statistical correctness. However, with the advent of increasingly large datasets and diverse sets of often complex and exploratory queries, computational considerations and automation (i.e., lack of reliance on deep analysis of the specific estimator and setting of interest) are increasingly important. Even as the amount of available data grows, the number of parameters to be estimated and the number of potential sources of bias often also grow, leading to a need to be able to tractably assess estimator quality in the setting of large data.

Thus, unlike previous work on estimator quality assessment, here we directly address computational costs and scalability in addition to statistical correctness and automation. Indeed, existing methods have significant drawbacks with respect to one or more of these desiderata. The bootstrap, despite its strong statistical properties, has high—even prohibitive—computational costs; thus, its usefulness is severely blunted by the large datasets increasingly encountered in practice. We also find that its relatives, such as the $m$ out of $n$ bootstrap and subsampling, can have lesser computational costs, as expected, but are generally not robust to specification of hyperparameters (such as the number of subsampled data points) and are also somewhat less automatic due to their need to explicitly utilize theoretically derived estimator convergence rates.

Motivated by the need for an automatic, accurate means of assessing estimator quality that is scalable to large datasets, we present a new procedure, the Bag of Little Bootstraps (BLB), which functions by combining the results of bootstrapping multiple small subsets of a larger original dataset. BLB has a significantly more favorable computational profile than the bootstrap, as it only requires repeated computation of the estimator under consideration on quantities of data that can be much smaller than the original dataset. Hence, BLB is well suited to implementation on modern distributed and parallel computing architectures. Our procedure maintains the bootstrap's automatic and generic applicability, favorable statistical properties, and simplicity of implementation. Finally, as we show empirically, BLB is consistently more robust than alternatives such as the $m$ out of $n$ bootstrap and subsampling.

We next formalize our statistical setting and notation in Section 2, discuss relevant prior work in Section 3, and present BLB in full detail in Section 4. Subsequently, we present an empirical and theoretical study of statistical correctness in Section 5, an exploration of scalability including large-scale experiments on a distributed computing platform in Section 6, practical methods for automatically selecting hyperparameters in Section 7, and assessments on real data in Section 8.

## 2. Setting and Notation

We assume that we observe a sample $X_1, \ldots, X_n \in \mathcal{X}$ drawn i.i.d. from some true (unknown) underlying distribution $P \in \mathcal{P}$. Based only on this observed data, we obtain an estimate $\hat{\theta}_n = \theta(\mathbb{P}_n) \in \Theta$, where

$\mathbb{P}_n = n^{-1} \sum_{i=1}^n \delta_{X_i}$ is the empirical distribution of $X_1, \ldots, X_n$. The true (unknown) population value to be estimated is $\theta(P)$. For example, $\hat{\theta}_n$ might estimate a measure of correlation, the parameters of a regressor, or the prediction accuracy of a trained classification model. Noting that $\hat{\theta}_n$ is a random quantity because it is based on $n$ random observations, we define $Q_n(P) \in \mathcal{Q}$ as the true underlying distribution of $\hat{\theta}_n$, which is determined by both $P$ and the form of the mapping $\theta$. Our end goal is the computation of some metric $\xi(Q_n(P)) : \mathcal{Q} \to \Xi$, for $\Xi$ a vector space, which informatively summarizes $Q_n(P)$. For instance, $\xi$ might compute a confidence region, a standard error, or a bias. In practice, we do not have direct knowledge of $P$ or $Q_n(P)$, and so we must estimate $\xi(Q_n(P))$ itself based only on the observed data and knowledge of the form of the mapping $\theta$.

## 3. Related Work

The bootstrap (Efron, 1979; Efron & Tibshirani, 1993) provides an automatic and widely applicable means of quantifying estimator quality: it simply uses the data-driven plugin approximation $\xi(Q_n(P)) \approx \xi(Q_n(\mathbb{P}_n))$. While $\xi(Q_n(\mathbb{P}_n))$ cannot be computed exactly in most cases, it is generally amenable to straightforward Monte Carlo approximation as follows: repeatedly resample $n$ points i.i.d. from $\mathbb{P}_n$, compute the estimate on each resample, form the empirical distribution $\mathbb{Q}_n$ of the computed estimates, and approximate $\xi(Q_n(P)) \approx \xi(\mathbb{Q}_n)$. Though conceptually simple and powerful, this procedure requires repeated estimator computation on resamples having size comparable to that of the original dataset. Therefore, if the original dataset is large, then this computation can be costly.

While the literature does contain some discussion of techniques for improving the computational efficiency of the bootstrap, that work is largely devoted to reducing the number of Monte Carlo resamples required (Efron, 1988; Efron & Tibshirani, 1993). These techniques in general introduce significant additional complexity of implementation and do not obviate the need for repeated estimator computation on resamples having size comparable to that of the original dataset.

Bootstrap variants such as the $m$ out of $n$ bootstrap (Bickel et al., 1997) and subsampling (Politis et al., 1999) were introduced to achieve statistical consistency in edge cases in which the bootstrap fails, though they also have the potential to yield computational benefits. The $m$ out of $n$ bootstrap (and subsampling) functions as follows, for $m < n$: repeatedly resample $m$ points i.i.d. from $\mathbb{P}_n$ (subsample $m$ points without replacement from $X_1, \ldots, X_n$), compute the estimate on each resample (subsample), form the empirical distribution $\mathbb{Q}_{n,m}$ of the computed estimates, approximate $\xi(Q_m(P)) \approx \xi(\mathbb{Q}_{n,m})$, and apply an analytical correction to in turn approximate $\xi(Q_n(P))$. This final analytical correction uses prior knowledge of the convergence rate of $\hat{\theta}_n$ as $n$ increases and is necessary because each estimate is computed based on only $m$ rather than $n$ points.

These procedures have a more favorable computational profile than the bootstrap, as they only require repeated estimator computation on smaller sets of data. However, they require knowledge and explicit use of the convergence rate of $\hat{\theta}_n$, and, as we show in our empirical investigation below, their success is sensitive to the choice of $m$. While schemes have been proposed for automatic selection of an optimal value of $m$ (Bickel & Sakov, 2008), they require significantly greater computation which would eliminate any computational gains. It is also worth noting that some work on the $m$ out of $n$ bootstrap has explicitly sought to reduce computational costs using two different values of $m$ in conjunction with extrapolation (Bickel & Yahav, 1988; Bickel & Sakov, 2002). However, these approaches explicitly use series expansions of the cdf values of the estimator's sampling distribution and hence are less automatically usable; they also require execution of the $m$ out of $n$ bootstrap for multiple different values of $m$.

## 4. Bag of Little Bootstraps (BLB)

The Bag of Little Bootstraps (Algorithm 1) functions by averaging the results of bootstrapping multiple small subsets of $X_1, \ldots, X_n$. More formally, given a subset size $b < n$, BLB samples $s$ subsets of size $b$ from the original $n$ data points, uniformly at random (one can also impose the constraint that the subsets be disjoint). Let $\mathcal{I}_1, \ldots, \mathcal{I}_s \subset \{1, \ldots, n\}$ be the corresponding index multisets (note that $|\mathcal{I}_j| = b, \forall j$), and let $\mathbb{P}_{n,b}^{(j)} = b^{-1} \sum_{i \in \mathcal{I}_j} \delta_{X_i}$ be the empirical distribution corresponding to subset $j$. BLB's estimate of $\xi(Q_n(P))$ is then given by $s^{-1} \sum_{j=1}^{s} \xi(Q_n(\mathbb{P}_{n,b}^{(j)}))$. Although the terms $\xi(Q_n(\mathbb{P}_{n,b}^{(j)}))$ cannot be computed analytically in general, they are computed numerically via the inner loop in Algorithm 1 via Monte Carlo approximation in the manner of the bootstrap: for each term $j$, we repeatedly resample $n$ points i.i.d. from $\mathbb{P}_{n,b}^{(j)}$, compute the estimate on each resample, form the empirical distribution $\mathbb{Q}_{n,j}^*$ of the computed estimates, and approximate $\xi(Q_n(\mathbb{P}_{n,b}^{(j)})) \approx \xi(\mathbb{Q}_{n,j}^*)$.

To realize the substantial computational benefits afforded by BLB, we utilize the following crucial fact: each BLB resample, despite having nominal size $n$,

---

**Algorithm 1** Bag of Little Bootstraps (BLB)

**Input:** Data $X_1, \ldots, X_n$
$\quad\quad\quad\theta$: estimator of interest
$\quad\quad\quad\xi$: estimator quality assessment
$\quad\quad\quad b$: subset size
$\quad\quad\quad s$: number of sampled subsets
$\quad\quad\quad r$: number of Monte Carlo iterations
**Output:** An estimate of $\xi(Q_n(P))$
**for** $j \leftarrow 1$ **to** $s$ **do**
$\quad$//Subsample the data
$\quad$Randomly sample a set $\mathcal{I} = \{i_1, \ldots, i_b\}$ of $b$ indices from $\{1, \ldots, n\}$ without replacement
$\quad$[or, choose $\mathcal{I}$ to be a disjoint subset of size $b$ from a predefined random partition of $\{1, \ldots, n\}$]
$\quad$//Approximate $\xi(Q_n(\mathbb{P}_{n,b}^{(j)}))$
$\quad$**for** $k \leftarrow 1$ **to** $r$ **do**
$\quad\quad$Sample $(n_1, \ldots, n_b) \sim \text{Multinomial}(n, \mathbf{1}_b/b)$
$\quad\quad\mathbb{P}_{n,k}^* \leftarrow n^{-1} \sum_{a=1}^{b} n_a \delta_{X_{i_a}}$
$\quad\quad\hat{\theta}_{n,k}^* \leftarrow \theta(\mathbb{P}_{n,k}^*)$
$\quad$**end for**
$\quad\mathbb{Q}_{n,j}^* \leftarrow r^{-1} \sum_{k=1}^{r} \delta_{\hat{\theta}_{n,k}^*}$
$\quad\xi_{n,j}^* \leftarrow \xi(\mathbb{Q}_{n,j}^*)$
**end for**
//Average values of $\xi(Q_n(\mathbb{P}_{n,b}^{(j)}))$ computed
//for different data subsets
**return** $s^{-1} \sum_{j=1}^{s} \xi_{n,j}^*$

---

contains at most $b$ distinct data points. In particular, to generate each resample, it suffices to draw a vector of counts from an $n$-trial uniform multinomial distribution over $b$ objects. We can then represent each resample by simply maintaining the at most $b$ distinct points present within it, accompanied by corresponding sampled counts (i.e., each resample requires only storage space in $O(b)$). In turn, if the estimator can work directly with this weighted data representation, then its computational requirements—with respect to both time and storage space—scale only in $b$, rather than $n$. Fortunately, this property does indeed hold for many if not most commonly used estimators, including M-estimators such as linear and kernel regression, logistic regression, and Support Vector Machines, among many others.

As a result, BLB only requires repeated computation on small subsets of the original dataset and avoids the bootstrap's problematic need for repeated computation of estimates on resamples having size comparable to that of the original dataset. A simple and standard calculation (Efron & Tibshirani, 1993) shows that each bootstrap resample contains approximately $0.632n$ distinct points, which is large if $n$ is large. In contrast, as discussed above, each BLB resample contains at most $b$ distinct points, and $b$ can be chosen to be much smaller than $n$ or $0.632n$. For example, we might take $b = n^\gamma$ where $\gamma \in [0.5, 1]$. More concretely, if $n = 1,000,000$, then each bootstrap resample would contain approximately 632,000 distinct points, whereas with $b = n^{0.6}$ each BLB subsample and resample would contain at most 3,981 distinct points. If each data point occupies 1 MB of storage space, then the original dataset would occupy 1 TB, a bootstrap resample would occupy approximately 632 GB, and each BLB subsample or resample would occupy at most 4 GB.

BLB thus has a significantly more favorable computational profile than the bootstrap. As seen in subsequent sections, our procedure typically requires less total computation to reach comparably high accuracy (fairly modest values of $s$ and $r$ suffice); is significantly more amenable to implementation on distributed and parallel computing architectures which are often used to process large datasets; maintains the favorable statistical properties of the bootstrap; and is more robust than the $m$ out of $n$ bootstrap and subsampling to the choice of subset size.

## 5. Statistical Correctness

We show via both a simulation study and theoretical analysis that BLB shares the favorable statistical performance of the bootstrap while also being consistently more robust than the $m$ out of $n$ bootstrap and subsampling to the choice of $b$. Here we present a representative summary of our investigation of statistical correctness; see Kleiner et al. (2012) for more detail.

We investigate the empirical performance characteristics of BLB and compare to existing methods via experiments on different simulated datasets and estimation tasks. Use of simulated data is necessary here because it allows knowledge of $Q_n(P)$ and hence $\xi(Q_n(P))$ (i.e., ground truth). We consider two different settings: regression and classification. For both settings, the data have the form $X_i = (\tilde{X}_i, Y_i) \sim P$, i.i.d. for $i = 1, \ldots, n$, where $\tilde{X}_i \in \mathbb{R}^d$; $Y_i \in \mathbb{R}$ for regression, whereas $Y_i \in \{0, 1\}$ for classification. We use $n = 20,000$ for the plots shown, and $d$ is set to 100 for regression and 10 for classification. In each case, $\hat{\theta}_n$ estimates a parameter vector in $\mathbb{R}^d$ (via either least squares or logistic regression with Newton's method, all implemented in MATLAB) for a linear or generalized linear model of the mapping between $\tilde{X}_i$ and $Y_i$. We define $\xi$ as computing a set of marginal 95% confidence intervals, one for each element of the estimated parameter vector (averaging across $\xi$'s consists of averaging element-wise interval boundaries).

To evaluate the various quality assessment procedures on a given estimation task and true underlying data distribution $P$, we first compute the ground truth $\xi(Q_n(P))$ based on 2,000 realizations of datasets of size $n$ from $P$. Then, for an independent dataset realization of size $n$ from the true underlying distribution, we run each quality assessment procedure and record the estimate of $\xi(Q_n(P))$ produced after each iteration (e.g., after each bootstrap resample or BLB subsample is processed), as well as the cumulative time required to produce that estimate. Every such estimate is evaluated based on the average (across dimensions) relative absolute deviation of its component-wise confidence intervals' widths from the corresponding true widths: given an estimated confidence interval width $c$ and a true width $c_o$, the relative deviation of $c$ from $c_o$ is defined as $|c - c_o|/c_o$. We repeat this process on five independent dataset realizations of size $n$ and average the resulting relative errors and corresponding times across these five datasets to obtain a trajectory of relative error versus time for each quality assessment procedure (the trajectories' variances are small relative to the relevant differences between their means, so the variances are not shown in our plots). To maintain consistency of notation, we henceforth refer to the $m$ out of $n$ bootstrap as the $b$ out of $n$ bootstrap. For BLB, the $b$ out of $n$ bootstrap, and subsampling, we consider $b = n^\gamma$ where $\gamma \in \{0.5, 0.6, 0.7, 0.8, 0.9\}$; we use $r = 100$ in all runs of BLB.

Figure 1 shows a set of representative results for the classification setting, where $P$ generates the components of $\tilde{X}_i$ i.i.d. from StudentT(3) and $Y_i \sim$ Bernoulli$((1 + \exp(-\tilde{X}_i^T \mathbf{1}))^{-1})$; we use this representative empirical setting in subsequent sections as well. As seen in the figure, BLB (left plot) succeeds in converging to low relative error more quickly than the bootstrap for $b > n^{0.5}$, while converging to somewhat higher relative error for $b = n^{0.5}$. We are more robust than the $b$ out of $n$ bootstrap (middle plot), which fails to converge to low relative error for $b \le n^{0.6}$. In fact, even for $b = n^{0.5}$, BLB's performance is substantially superior to that of the $b$ out of $n$ bootstrap. For the aforementioned case in which BLB does not match the relative error of the bootstrap, additional empirical results (right plot) and our theoretical analysis indicate that this discrepancy in relative error diminishes as $n$ increases. Identical evaluation of subsampling (plots not shown) shows that it performs strictly worse than the $b$ out of $n$ bootstrap. Qualitatively similar results also hold in both the classification and regression settings (with the latter generally showing better performance) when $P$ generates $\tilde{X}_i$ from Normal, Gamma, or StudentT distributions, and when $P$ uses a non-linear noisy mapping between $\tilde{X}_i$ and $Y_i$ (so that we estimate a misspecified model).

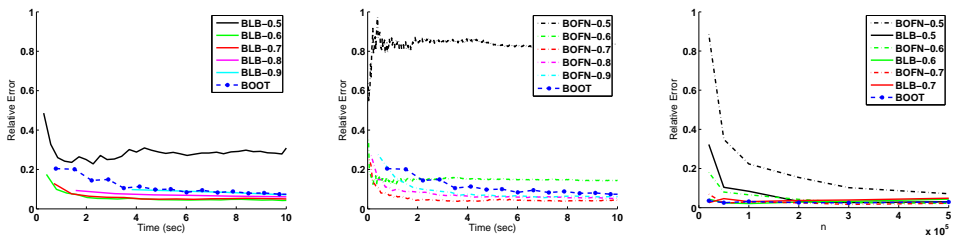

*Figure 1.* Results for classification setting with linear data generating distribution and StudentT $\tilde{X}_i$ distribution. For both BLB and the $b$ out of $n$ bootstrap (BOFN), $b = n^\gamma$ with the value of $\gamma$ for each trajectory given in the legend. **(left and middle)** Relative error vs. processing time, with $n = 20,000$. The left plot shows BLB with bootstrap (BOOT), and the middle plot shows BOFN. **(right)** Relative error (after convergence) vs. $n$ for BLB, BOFN, and BOOT.

These experiments illustrate the statistical correctness of BLB, as well as its improved robustness to the choice of $b$. The results are borne out by our theoretical analysis, which shows that BLB has statistical properties that are identical to those of the bootstrap, under the same conditions that have been used in prior analysis of the bootstrap. In particular, BLB is asymptotically consistent for a broad class of estimators and quality measures (see Kleiner et al. (2012) for proof):

**Theorem 1.** *Suppose that $\theta$ is Hadamard differentiable at $P$ tangentially to some subspace, with $P$ and $\mathbb{P}_{n,b}^{(j)}$ viewed as maps from some Donsker class $\mathcal{F}$ to $\mathbb{R}$ such that $\mathcal{F}_\delta$ is measurable for every $\delta > 0$, where $\mathcal{F}_\delta = \{f - g : f, g \in \mathcal{F}, \rho_P(f - g) < \delta\}$ and $\rho_P(f) = (P(f - Pf)^2)^{1/2}$. Additionally, assume that $\xi$ is continuous in the space of distributions $\mathcal{Q}$ with respect to a metric that metrizes weak convergence. Then, up to centering and rescaling of $\hat{\theta}_n$, $s^{-1}\sum_{j=1}^{s} \xi(Q_n(\mathbb{P}_{n,b}^{(j)})) - \xi(Q_n(P)) \xrightarrow{P} 0$ as $n \to \infty$, for any sequence $b \to \infty$ and for any fixed $s$. Furthermore, the result holds for sequences $s \to \infty$ if $E|\xi(Q_n(\mathbb{P}_{n,b}^{(j)}))| < \infty$.*

BLB is also higher-order correct: despite the fact that the procedure only applies the estimator in question to subsets of the full observed dataset, it shares the fast convergence rates of the bootstrap, which permit convergence of the procedure's output at rate $O(1/n)$ rather than the rate of $O(1/\sqrt{n})$ achieved by asymptotic approximations. To achieve these fast convergence rates, some natural conditions on BLB's hyperparameters are required, for example that $b = \Omega(\sqrt{n})$ and that $s$ be sufficiently large with respect to the variability in the observed data. Quite interestingly, these conditions permit $b$ to be significantly smaller than $n$, with $b/n \to 0$ as $n \to \infty$. See Kleiner et al. (2012) for our theoretical results on higher-order correctness.

## 6. Scalability

The experiments of the preceding section, though primarily intended to investigate statistical performance, also provide some insight into computational performance: as seen in Figure 1, when computing serially, BLB generally requires less time, and hence less total computation, than the bootstrap to attain comparably high accuracy. Those results only hint at BLB's superior ability to scale computationally to large datasets, which we now demonstrate in full in the following discussion and via large-scale experiments.

Modern massive datasets often exceed both the processing and storage capabilities of individual processors or compute nodes, thus necessitating the use of parallel and distributed computing architectures. Indeed, in the large data setting, computing a single full-data point estimate often requires simultaneous distributed computation across multiple compute nodes, among which the observed dataset is partitioned. As a result, the scalability of a quality assessment method is closely tied to its ability to effectively utilize such computing resources.

Due to the large size of bootstrap resamples (recall that approximately 63% of data points appear at least once in each resample), the following is the most natural avenue for applying the bootstrap to large-scale data using distributed computing: given data on a cluster of compute nodes, parallelize the estimate computation on each resample across the cluster, and compute on one resample at a time. This approach, while at least potentially feasible, remains quite problematic. Each estimate computation requires the use of an entire cluster of compute nodes, and the bootstrap repeatedly incurs the associated overhead, such as the cost of repeatedly communicating intermediate data among nodes. Additionally, many cluster computing systems in widespread use (e.g., Hadoop MapReduce) store data only on disk, rather than in memory, due to physical size constraints (if the data size exceeds the amount of available memory) or architectural constraints (e.g., the need for fault tolerance). In that case, the bootstrap incurs the extreme costs associated with repeatedly reading a very large dataset from disk; though disk read costs may be acceptable when (slowly) computing only a single point estimate, they easily become prohibitive when computing many estimates on one hundred or more resamples.

In contrast, BLB permits computation on multiple (or even all) subsamples and resamples simultaneously in parallel, allowing for straightforward and effective distributed and parallel implementations which enable effective scalability and large computational gains. Because BLB subsamples and resamples can be significantly smaller than the original dataset, they can be transferred to, stored by, and processed on individual (or very small sets of) compute nodes. For example, we can naturally leverage modern hierarchical distributed architectures by distributing subsamples to different compute nodes and subsequently using intranode parallelism to compute across different resamples generated from the same subsample. Note that generation and distribution of the subsamples requires only a single pass over the full dataset (i.e., only a single read of the full dataset from disk, if it is stored only on disk), after which all required data (i.e., the subsamples) can be stored in memory. Beyond this significant architectural benefit, we also achieve implementation and algorithmic benefits: we do not need to parallelize the estimator computation internally, as BLB uses the available parallelism to compute on multiple resamples simultaneously, and exposing the estimator to only $b$ rather than $n$ distinct points significantly reduces the computational cost of estimation, particularly if the estimator computation scales super-linearly.

We now empirically substantiate the preceding discussion via large-scale experiments performed on Amazon EC2. We use the representative experimental setup of Figure 1, but now with $d = 3,000$, $n = 6,000,000$, $Y_i \sim \text{Bernoulli}((1 + \exp(-\tilde{X}_i^T \mathbf{1}/\sqrt{d}))^{-1})$, and the logistic regression implemented using L-BFGS. The size of a full observed dataset in this setting is thus approximately 150 GB. We compare the performance of BLB and the bootstrap (now omitting the $m$ out of $n$ bootstrap and subsampling due to the shortcomings illustrated in Section 5), both implemented as described above (we parallelize the estimator computation for the bootstrap by simply distributing gradient computations via MapReduce) using the Spark cluster computing framework (Spark, 2012), which provides the ability to either read data from disk or cache it in memory (provided that sufficient memory is available) for faster repeated access. For BLB, we use $r = 50$, $s = 5$, and $b = n^{0.7}$. Due to the larger data size and use of a distributed architecture, we now implement the bootstrap using Poisson resampling, and we compute ground truth using 200 independent dataset realizations.

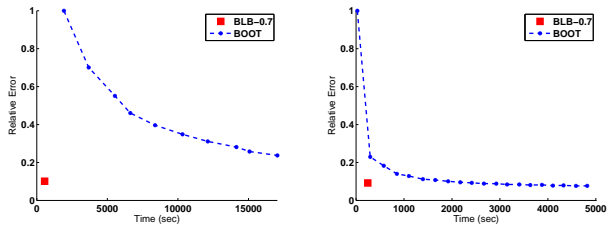

*Figure 2.* Large-scale results for BLB and the bootstrap (BOOT) on 150 GB of data. Because BLB is fully parallelized across subsamples, we only show the relative error and total processing time of its final output. **(left)** Data stored on disk. **(right)** Data cached in memory.

In the left plot of Figure 2, we show results obtained using a cluster of 10 worker nodes, each having 6 GB of memory and 8 compute cores; thus, the total memory of the cluster is 60 GB, and the full dataset (150 GB) can only be stored on disk. As expected, the time required by the bootstrap to produce even a low-accuracy output is prohibitively high, while BLB provides a high-accuracy output quite quickly, in less than the time required to process even a single bootstrap resample. In the right plot of Figure 2, we show results obtained using a cluster of 20 worker nodes, each having 12 GB of memory and 4 compute cores; thus, the total memory of the cluster is 240 GB, and we cache the full dataset in memory for fast repeated access. Unsurprisingly, the bootstrap's performance improves significantly with respect to the previous experiment. However, the performance of BLB (which also improves), remains substantially better than that of the bootstrap.

Thus, relative to the bootstrap, BLB both allows more natural and effective use of parallel and distributed computational resources and decreases the total computational cost of assessing estimator quality. Finally, it is worth noting that even if only a single compute node is available, BLB allows the following somewhat counterintuitive possibility: even if it is prohibitive to actually compute a point estimate for the full observed data using a single compute node (because the full dataset is large), it may still be possible to efficiently assess such a point estimate's quality using only a single compute node by processing one subsample (and the associated resamples) at a time.

## 7. Hyperparameter Selection

Like existing resampling-based procedures such as the bootstrap, BLB requires the specification of hyperparameters controlling the number of subsamples and resamples processed. Setting such hyperparameters to be sufficiently large is necessary to ensure good statistical performance; however, setting them to be unnecessarily large results in wasted computation. Prior work on the bootstrap and related procedures generally assumes that a procedure's user will simply select a priori a large, constant number of resamples to be processed (with the exception of Tibshirani (1985), who does not provide a general solution to this issue). However, this approach reduces the level of automation of these methods and can be quite inefficient in the large data setting.

Thus, we now examine the dependence of BLB's performance on the choice of $r$ and $s$, with the goal of better understanding their influence and providing guidance and adaptive (i.e., more automatic) methods for their selection. The left plot of Figure 3 illustrates the influence of $r$ and $s$, giving the relative errors achieved by BLB with $b = n^{0.7}$ for different $r, s$ pairs in the representative empirical setting described in Section 5. In particular, note that for all but the smallest values of $r$ and $s$, it is possible to choose these values independently such that BLB achieves low relative error; in this case, choosing $s \geq 3, r \geq 50$ is sufficient.

While these results are useful and provide some guidance, we expect the minimal sufficient values of $r$ and $s$ to change based on the identity of $\xi$ (e.g., we expect a confidence interval to be harder to compute and hence to require larger $r$ than a standard error) and the properties of the underlying data. Thus, to help avoid the need to choose $r$ and $s$ conservatively, we now provide a means for adaptive hyperparameter selection, which we validate empirically.

Concretely, to select $r$ adaptively in the inner loop of Algorithm 1, we propose, for each subsample $j$, to continue to process resamples and update $\xi_{n,j}^*$ until it has ceased to change significantly. Noting that the values $\hat{\theta}_{n,k}^*$ used to compute $\xi_{n,j}^*$ are conditionally i.i.d. given a subsample, for most forms of $\xi$ the series of computed $\xi_{n,j}^*$ values will be well behaved and will converge (in many cases at rate $O(1/\sqrt{r})$, though with unknown constant) to a constant target value as more resamples are processed. Therefore, it suffices to process resamples (i.e., to increase $r$) until we are satisfied that $\xi_{n,j}^*$ has ceased to fluctuate significantly; we propose using Algorithm 2 to assess this convergence. The same scheme can be used to select $s$ adaptively by processing more subsamples (i.e., increasing $s$) until BLB's output value $s^{-1} \sum_{j=1}^s \xi_{n,j}^*$ has stabilized; in this case, one can simultaneously also choose $r$ adaptively and independently for each subsample.

The middle plot of Figure 3 shows the results of applying such adaptive hyperparameter selection. For selection of $r$ we use $\epsilon = 0.05$ and $w = 20$, and for selection of $s$ we use $\epsilon = 0.05$ and $w = 3$. As seen in the plot, the adaptive hyperparameter selection allows BLB to cease computing shortly after it has converged (to low relative error), limiting the amount of unnecessary computation that is performed. Though selected a priori, $\epsilon$ and $w$ are more intuitively interpretable and less dependent on the details of $\xi$ and the underlying data generating distribution than $r$ and $s$. Indeed, the aforementioned specific values of $\epsilon$ and $w$ yield results of comparably good quality when also used for a variety of other synthetic and real data generation settings (see Section 8 below), as well as for different forms of $\xi$ (see the righthand table in Figure 3, which shows that smaller values of $r$ are selected when $\xi$ is easier to compute). Thus, our scheme significantly helps to alleviate the burden of hyperparameter selection.

Automatic selection of a value of $b$ in a computationally efficient manner is more difficult due to the inability to reuse computations performed for different values of $b$. One could consider similarly increasing $b$ from some small value until the output of BLB stabilizes; devising a means of doing so efficiently is the subject of future work. Nonetheless, based on our fairly extensive empirical investigation, it seems that $b = n^{0.7}$ is a reasonable and effective choice in many situations.

## 8. Real Data

We now present the results of applying BLB to several different real datasets. In this setting, given the absence of ground truth, it is not possible to objectively evaluate estimator quality assessment methods' statistical correctness. As a result, we are reduced to comparing the outputs of different methods to each other; we also now report the average (across dimensions) absolute confidence interval width produced by each procedure, rather than relative error.

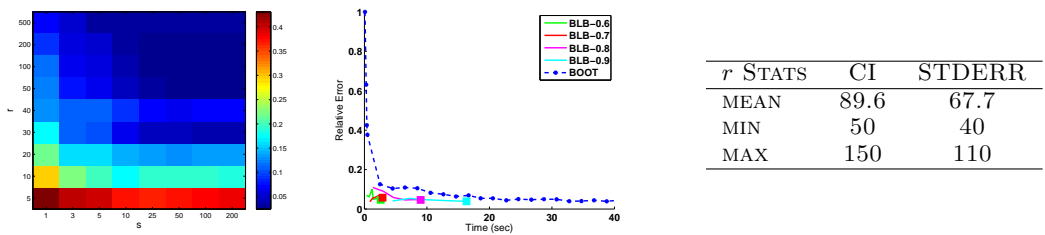

| $r$ Stats | CI | STDERR |
|-----------|------|--------|
| MEAN | 89.6 | 67.7 |
| MIN | 50 | 40 |
| MAX | 150 | 110 |

*Figure 3.* Results for BLB hyperparameter selection in the empirical setting of Figure 1, with $b = n^{0.7}$. **(left)** Relative error achieved by BLB for different values of $r$ and $s$. **(middle)** Relative error vs. processing time (without parallelization) for BLB using adaptive selection of $r$ and $s$ (resulting stopping times of BLB trajectories are marked by squares) and the bootstrap (BOOT). **(right)** Statistics of the different values of $r$ selected by BLB's adaptive hyperparameter selection (across multiple subsamples) when $\xi$ is either our usual confidence interval-based quality measure (CI), or a component-wise standard error (STDERR); the relative errors achieved by BLB and the bootstrap are comparable in both cases.

Figure 4 shows results for BLB, the bootstrap, and the $b$ out of $n$ bootstrap on the UCI connect4 dataset (logistic regression, $d = 42$, $n = 67,557$). We select the BLB hyperparameters $r$ and $s$ using the adaptive method described in the previous section. Notably, the outputs of BLB for all values of $b$ considered, and the output of the bootstrap, are tightly clustered around the same value; additionally, as expected, BLB converges more quickly than the bootstrap. However, the values produced by the $b$ out of $n$ bootstrap vary significantly as $b$ changes, thus further highlighting this procedure's lack of robustness. We have obtained qualitatively similar results on six additional UCI datasets (ct-slice, magic, millionsong, parkinsons, poker, shuttle) with different estimators (linear regression and logistic regression) and a range of different values of $n$ and $d$; see Kleiner et al. (2012) for additional plots.

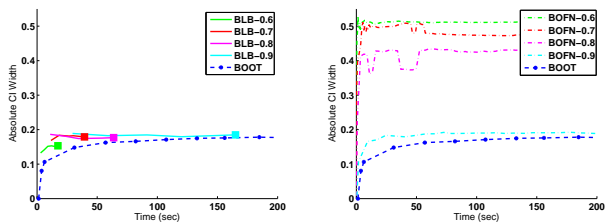

*Figure 4.* Average absolute confidence interval width vs. processing time on the UCI connect4 dataset (logistic regression, $d = 42, n = 67,557$). **(left)** Results for BLB (with adaptive hyperparameter selection) and bootstrap (BOOT). **(right)** Results for $b$ out of $n$ bootstrap (BOFN).

## 9. Conclusion

We have presented a new procedure, BLB, which provides a powerful new alternative for automatic, accurate assessment of estimator quality that is well suited to large-scale data and modern parallel and distributed computing architectures.

## Acknowledgments

This research is supported in part by an NSF CISE Expeditions award, gifts from Google, SAP, Amazon Web Services, Blue Goji, Cisco, Cloudera, Ericsson, General Electric, Hewlett Packard, Huawei, Intel, MarkLogic, Microsoft, NetApp, Oracle, Quanta, Splunk, VMware and by DARPA (contract #FA8650-11-C-7136). Ameet Talwalkar was supported by NSF award No. 1122732.